\documentclass[letterpaper]{article}
\usepackage{aaai2027}
\usepackage[hyphens]{url}
\usepackage{graphicx}
\usepackage{natbib}
\usepackage{bibunits}
\usepackage{caption}
\usepackage{subcaption}
\usepackage{float}
\usepackage{booktabs}
\usepackage{multirow}
\usepackage{array}
\usepackage{amsmath}
\usepackage{amssymb}
\newcommand{\ms}[2]{#1{\scriptstyle\,\pm\,#2}}
\defaultbibliography{references}
\defaultbibliographystyle{aaai2027}

\title{Does Latent Context Help? A Controlled Evaluation of\\Inverse Reinforcement Learning in Arctic Shipping}
\author{
Vaishnav Vaidheeswaran,\textsuperscript{\rm 1}
Dilith Jayakody,\textsuperscript{\rm 1}
Biruk Ambaw,\textsuperscript{\rm 1}\\
Jaswanth Kumar,\textsuperscript{\rm 2}
Md Mahbub Alam,\textsuperscript{\rm 1}
Gabriel Spadon\textsuperscript{\rm 1,*}
}
\affiliations{
\textsuperscript{\rm 1}Faculty of Computer Science, Dalhousie University, Halifax, NS, Canada\\
\textsuperscript{\rm 2}Department of Industrial Engineering, Faculty of Engineering, Dalhousie University, Halifax, NS, Canada\\
\textsuperscript{*}Corresponding author: spadon@dal.ca
}
\begin{document}

\maketitle

\begin{bibunit}

\begin{abstract}
Artificial Intelligence (AI)-assisted navigation can help Arctic shipping adapt to rapidly changing sea-ice conditions, but reliable deployment requires reward models that are interpretable and robust to changing environments. Inverse reinforcement learning (IRL) provides a framework for recovering such rewards from vessel trajectories, while recent meta-IRL methods introduce latent context variables to capture behavioral heterogeneity. However, it remains unclear whether these latent representations recover genuinely hidden preferences or simply re-encode information already available in the observed state. We conduct a controlled evaluation on 3,186 AIS-derived voyages from 202 vessels across nine Arctic shipping seasons, comparing a linear shared reward, a nonlinear shared reward, and a latent-context model built on the same nonlinear architecture. The nonlinear reward improves held-out likelihood by 50.9\% over the linear baseline, whereas adding vessel-specific latent context reduces performance by 16.5\%. Behavioral analysis, context probes, and a pre-registered feature-hiding ablation show that apparent vessel-level variation is largely explained by observable route and environmental conditions rather than hidden vessel-specific factors. Moreover, predictive accuracy, route fidelity, and reward transfer yield different model rankings, demonstrating that no single metric is sufficient to evaluate learned rewards. These findings motivate testing whether the observed route, environmental, and vessel features already explain behavioral variation before adding per-vessel latent context. This supports more trustworthy AI deployment in safety-critical domains.
\end{abstract}

\section{Introduction}
\label{sec:intro}
\vspace{-2pt}
Arctic sea ice is retreating faster than navigational experience is
being acquired. Shipping seasons are lengthening, corridors that were
impassable a decade ago now carry commercial traffic, and an ice regime
that made a route safe in one season can make the same route hazardous in
the next \cite{shu2026integrating}.
Such expertise is difficult to transfer because it is held by a small number of ice navigators, depends on changing conditions, and is not recorded in a readily reusable form.

Learning from demonstrations provides a way to capture such expertise,
but the representation learned from demonstrations is critical. Behavioral
cloning can reproduce historical decisions but may fail when conditions
change, whereas inverse reinforcement learning (IRL) recovers the
underlying objectives that guide decisions and allows behavior to be
re-optimized under new conditions \cite{ziebart2008maximum,ziebart2010modeling}.
Because Arctic sea-navigation conditions vary across seasons and years,
transferable reward models provide a more adaptable alternative to
memorizing past behavior.

Real-world demonstrations, however, rarely come from identical decision
makers. Vessels differ in capability, cargo, schedule pressure, and risk
tolerance. Meta-IRL methods address this heterogeneity by introducing
latent context variables that condition rewards on demonstrator-specific
preferences \cite{yu2019meta,li2017infogail,xu2019learning}. While this
assumption is well motivated, it has primarily been studied on synthetic
benchmarks where task variation is predefined. In real environments,
behavioral differences may instead arise from hidden preferences,
observable conditions, or both. Thus, behavioral variation alone does not
establish the need for latent context.

This question is important for safety-relevant maritime applications,
where learned behavioral models can support route analysis, situational
awareness, and decision support under changing environmental conditions.
Our goal is not to deploy autonomous navigation, but to evaluate whether
learned rewards provide reliable representations of observed behavior.

We investigate this question through a controlled study of nine seasons
of Arctic Automatic Identification System (AIS) traffic, comprising
$3{,}186$ voyages from $202$ cargo and tanker vessels. We represent
navigation as a hexagonal graph Markov decision process (MDP) and examine
whether vessel-level behavioral variation requires latent context or can
be explained by observable features.

We compare three reward models that isolate reward flexibility from latent entity-specific adaptation (hereafter, personalization): MCE-IRL \cite{ziebart2010modeling}, which learns a shared linear reward; AIRL \cite{fu2018learning}, which introduces nonlinear shared rewards; and PEMIRL \cite{yu2019meta}, which adds per-vessel latent context while retaining nonlinear rewards. Here, personalization refers to learning latent vessel-specific representations that adapt decision-making behavior based on information not captured by the observable state, rather than user preference modeling or recommendation.

We make three contributions. First, we provide a controlled empirical evaluation of latent-context IRL on large-scale real-world trajectories, separating the effects of reward flexibility from latent context. Second, we develop analyses that distinguish behavioral variation explained by observable route, environmental, and vessel factors from unexplained variation. Third, we introduce a context-need diagnostic for determining when per-vessel latent context is likely to add decision-relevant information beyond the observed state. Our findings show that behavioral variation alone is insufficient evidence for latent context and motivate more careful evaluation of entity-specific adaptive AI systems.

\section{Related Work}
\label{sec:related}
\vspace{-2pt}
Classical maximum-entropy inverse reinforcement learning (MaxEnt IRL) learns reward functions from expert demonstrations under a maximum-entropy objective \cite{ziebart2008maximum,ziebart2010modeling}. In our experiments, maximum causal entropy IRL (MCE-IRL) represents one shared reward as a linear combination of observed features. Later methods introduced nonlinear reward representations. Guided Cost Learning
estimates a neural cost using sampled trajectories \cite{finn2016guided}, while Generative Adversarial Imitation Learning (GAIL) learns a policy through a discriminator that distinguishes expert from generated behavior \cite{ho2016generative, osa2018algorithmic}. Adversarial Inverse Reinforcement Learning (AIRL) \cite{fu2018learning} adapts this framework for reward recovery by separating the underlying task objective from a potential-based shaping term, to improve transfer beyond training dynamics. We use MCE-IRL and AIRL as shared-reward models: both assume that one reward explains all demonstrators, but AIRL can represent nonlinear
interactions among observable route, vessel, and environmental features.

When demonstrations may arise from different tasks, goals, or behavioral modes, one approach is to infer a latent context from trajectories and condition the model on it. InfoGAIL \cite{li2017infogail} uses discrete latent codes to identify imitation styles, whereas Probabilistic Embeddings for Meta-IRL (PEMIRL) \cite{yu2019meta} infers a distribution over task contexts and conditions both the reward $f(s,a,z)$ and policy on the latent variable $z$. PEMIRL uses an information-maximization objective, similar to InfoGAN \cite{chen2016infogan}, to prevent the model from ignoring the latent context. Other meta-IRL approaches use gradient-based meta-learning to adapt reward parameters from a small number of demonstrations \cite{finn2017model, xu2019learning}. Probabilistic context inference is also used in meta-RL, where recent experience identifies the current task and conditions the policy~\cite{rakelly2019efficient}.

These methods are commonly evaluated on benchmarks where meaningful task
differences are present by construction. Such benchmarks test whether a
model can recover latent variation known to exist, but not whether a
latent variable is needed for naturally occurring behavioral
differences. In real-world data, apparent differences may arise from
hidden preferences or from observable environmental conditions, route
assignments, and physical characteristics. We therefore treat vessel
identity as a candidate source of latent context, rather than evidence
that vessels optimize different hidden objectives, and ask whether
vessel-specific variation remains after observable route, sea-ice,
environmental, and vessel features are considered. This mirrors a long-standing distinction in econometric and transportation modeling between heterogeneity captured by observed covariates and genuinely unobserved heterogeneity requiring random or latent parameters \cite{heckman1981, mannering2016}; the latent context in Eq.~\ref{eq:infomax} is the IRL analogue of a random-parameters term, and the question is whether the data warrant one.

Accurate prediction of held-out expert actions does not necessarily
imply that a learned reward is useful to optimize. An imperfect proxy
may fit demonstrations while assigning high reward to unintended
behavior that a new agent can exploit~\cite{amodei2016concrete, skalse2022defining}. Studies of adversarial
imitation learning also show that implementation, regularization,
normalization, demonstration quality, and optimization choices can
affect performance as much as the nominal algorithm
\cite{orsini2021matters}. We therefore hold data, features, splits,
optimization budget, and random seeds fixed while changing one modeling
factor at a time. Each reward is evaluated by its held-out prediction,
complete-route fidelity, and ability to train a newly initialized policy.

Related navigation work has produced several models. Learned
costs support personalized A* route recommendation~\cite{wang2019empowering}, while trajectory-prediction and mining methods forecast movement or identify mobility patterns~\cite{nguyen2024traisformer, spadon2025modeling, zheng2015trajectory}.
In maritime settings, prior studies have optimized routes
using fuel and risk objectives, trained goal-conditioned
policies with hand-designed rewards, and assessed operations under
changing sea-ice conditions~\cite{zhang2022three, vaidheeswaran2025goal, shu2026integrating}.
These studies predict movement, characterize traffic, optimize routes,
learn policies, or assess risk. To the best of our knowledge, prior work has not
isolated whether a reward recovered from vessel trajectories
benefits from vessel-specific latent context after controlling for
nonlinear model capacity and observable route and environmental
information.

\section{Methodology}

\subsection{Preliminaries}
\label{sec:background}
\vspace{-2pt}
We work in a deterministic, goal-conditioned MDP
$\mathcal{M}=(\mathcal{S},\mathcal{A},T,\phi,M,\gamma,H)$, where states
are navigable cells, $T(s,a)$ is the deterministic successor, and
$\phi(s)\in\mathbb{R}^D$ is a shared feature vector. A validity mask
$M(s,a)$ marks which actions exist at $s$; invalid actions receive zero
probability in every policy, likelihood, chance baseline, and decoded
route. A demonstration $\tau=(s_0,a_0,\dots,s_L)$ is one voyage with
demonstrator index $v$ and absorbing goal $g$.

MCE-IRL \cite{ziebart2010modeling} assumes a linear reward
$r_\theta(s)=\theta^\top\phi(s)$ and the maximum-causal-entropy policy it
induces. Because our graph diameter ($\approx250$ cells) requires a
finite horizon, soft value iteration proceeds for
$t=H-1,\dots,0$ as
\begin{equation}
\begin{aligned}
Q_t(s,a)&=r_\theta(s)+V_{t+1}(T(s,a)),\\
V_t(s)&=\log\!\sum_{a\in\mathcal A(s)}
        \exp(Q_t(s,a)).
\end{aligned}
\label{eq:softvi}
\end{equation}

with $V_t(g)=0$ and
$\log\pi_t(a\mid s)=Q_t(s,a)-V_t(s)$. Maximizing demonstration
log-likelihood is convex in $\theta$, with gradient equal to the
difference between expert and model feature expectations. The resulting
policy is non-stationary because it depends on the remaining horizon.

AIRL \cite{fu2018learning} replaces the linear reward with a learned
nonlinear reward. In the PEMIRL implementation, the discriminator
predicts expert transitions and its logit $f(s,a)$ is used directly as
the reward; the generator is trained by PPO
\cite{schulman2017proximal}. Canonical AIRL instead uses
\[
D(s,a,s')=\frac{\exp(f(s,a,s'))}
{\exp(f(s,a,s'))+\pi(a\mid s)},
\]
with
\begin{equation}
f(s,s')=g(s)+\gamma h(s')-h(s),
\label{eq:airl}
\end{equation}
where the state-only $g$ is the transferable reward and $h$ is a shaping
term. We evaluate both formulations.

PEMIRL \cite{yu2019meta} extends this framework by conditioning both
reward and policy on a latent context. A posterior
$q(z\mid\tau^{1:k})$ infers $z\in\mathbb{R}^{d_z}$ from a
demonstrator's support trajectories, the reward becomes $f(s,a,z)$, and
the policy becomes $\pi(a\mid s,z)$. To prevent the latent context from
being ignored, PEMIRL adds a reward-weighted information-maximization
term inspired by InfoGAN \cite{chen2016infogan}:
\begin{equation}
\mathcal{L}_{\mathrm{info}}
=
-\eta\,
\mathbb{E}\!\left[
\log q(z\mid\tau)
\left(
\sum_t f(s_t,a_t,z)-\bar f
\right)
\right],
\label{eq:infomax}
\end{equation}
where $\bar f$ is the mean trajectory reward within a rollout group.
AIRL is the $z$-free special case of the same model, making the pair a
clean control: only the context pathway differs. Reporting only
MCE-IRL against PEMIRL, the usual evaluation, confounds nonlinear reward
capacity with latent-context sharing; its outcome is the sum of two
effects that, as we show, point in opposite directions.

\begin{figure*}[t]
\centering

\begin{subfigure}[t]{0.7\textwidth}
    \centering
    \includegraphics[width=\linewidth]{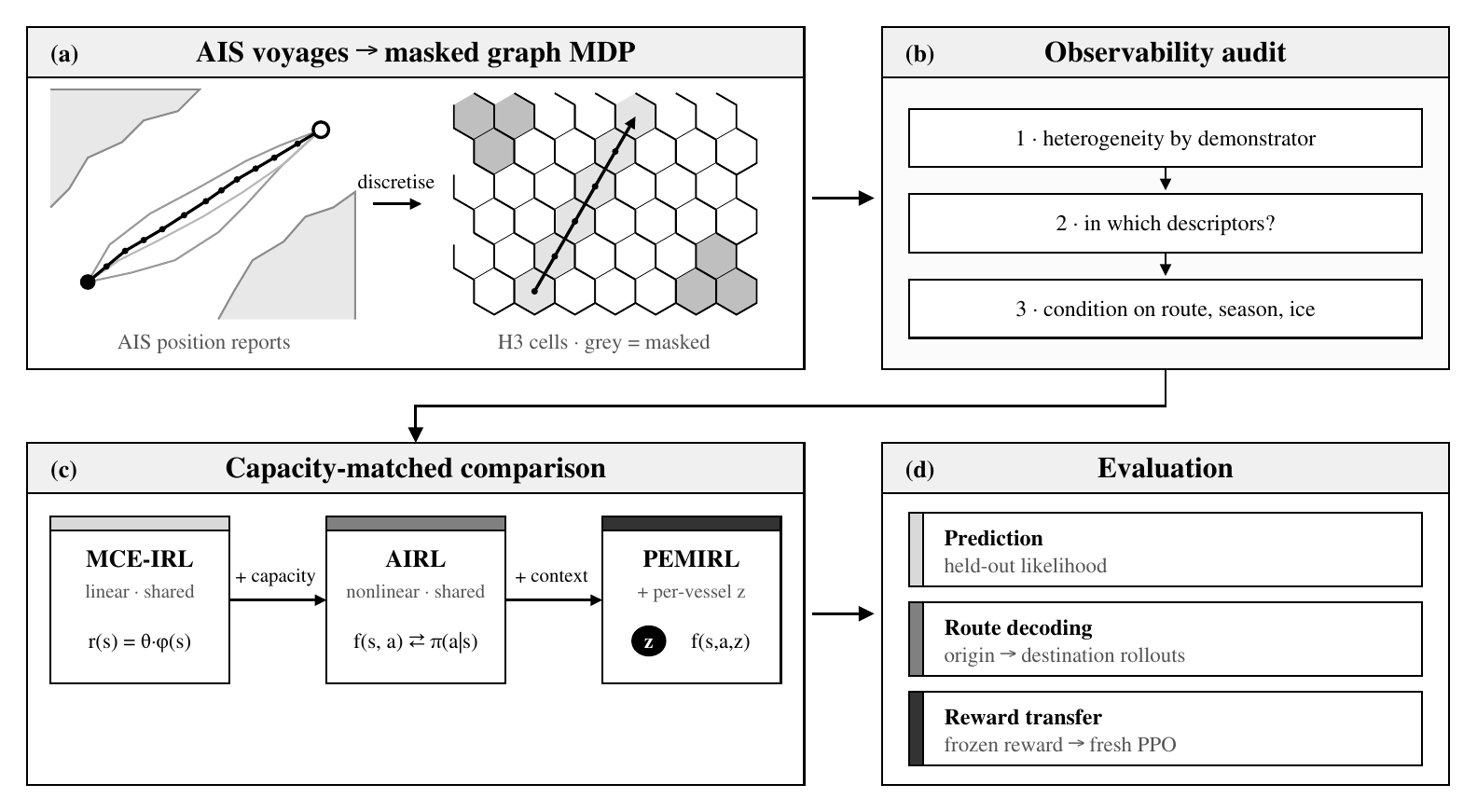}
    \caption{}
    \label{fig:arch}
\end{subfigure}
\hfill
\begin{subfigure}[t]{0.28\textwidth}
    \centering
    \includegraphics[width=\linewidth]{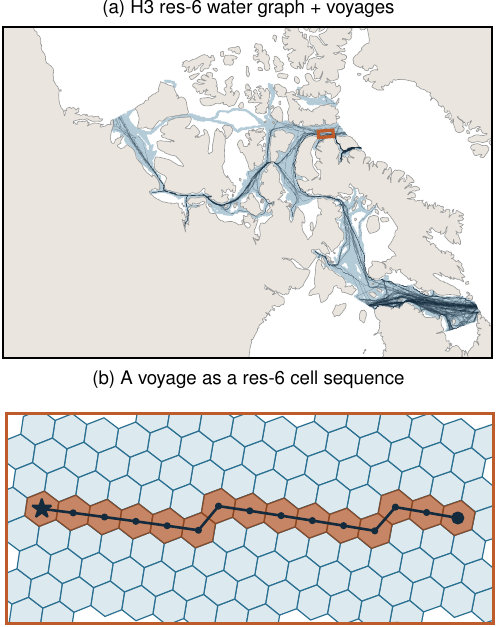}
    \caption{}
    \label{fig:hexgrid}
\end{subfigure}

\caption{Overview of the proposed framework and study region. (i) End-to-end
workflow of the study. Raw AIS trajectories are represented as a masked graph
MDP; state features are selectively hidden to create partially observable
settings; and reward functions learned under full and partial observability
are compared using reward recovery, trajectory fidelity, and
out-of-distribution generalization. (ii) Study area with AIS trajectories
discretized onto the H3 hexagonal grid.}
\label{fig:overview}
\end{figure*}

Figure~\ref{fig:overview}(i) summarizes the methodology. We convert AIS voyages into demonstrations on a masked hexagonal graph MDP. We then test whether
vessel differences remain after conditioning on observable factors.
Finally, we compare reward models that isolate reward expressiveness
(referred to as reward capacity) and latent context. We evaluate them
using held-out prediction, route fidelity, and their ability to train a
new policy.

\subsection{Trajectory Corpus and Preprocessing}
\vspace{-2pt}
We use AIS data for cargo and tanker vessels operating in the Canadian
Arctic during the July--October seasons of 2016--2024. We clean the
positional streams and segment them into individual voyages, which
serve as expert demonstrations. Following \citet{spadon2025modeling},
we discretize each trajectory onto the H3\footnote{An open-source
hierarchical hexagonal geospatial index; see \url{https://h3geo.org/}.}
hexagonal spatial index at resolution~6, where each cell is
approximately $6.9$\,km wide. Figure~\ref{fig:overview}(ii) shows AIS
voyages after discretization onto the H3 grid. We collapse repeated cells, bridge
single-cell gaps, discard voyages with fewer than five transitions, and
cap episode length at $H=512$ steps. The resulting navigation graph
contains $|\mathcal{S}|=14{,}206$ water states and $39{,}053$ edges.
The final dataset contains $3{,}186$ voyage demonstrations from $202$
vessels: $2{,}375$ cargo voyages and $811$ tanker voyages.

\subsection{Navigation MDP and State Representation}
\vspace{-2pt}
We model each voyage as a goal-conditioned MDP
\cite{liu2022goal}. Each episode has a start cell and a goal cell, and
the vessel moves between adjacent H3 cells. The action mask
$M(s,a)$~\cite{huang2020closer} removes unavailable moves at coastal
and pentagon cells.

Our environment follows the graph-navigation setting of
\citet{vaidheeswaran2025goal}. Their agent selects both direction and
speed, whereas we model route direction only because speed does not
change graph transitions. The action space has $|\mathcal{A}|=7$ actions. Action $0$ keeps the
vessel in its current cell, and actions $1$--$6$ move to neighboring
cells in clockwise bearing order from north. At cells with fewer than
six valid neighbors, $M(s, a)$ masks unavailable actions by setting their
logits to $-10^9$. We recover actions from consecutive cell pairs, so
the demonstrations and trained models use the same MDP. We set
$\gamma=0.99$.

All models use $\phi(s)\in\mathbb{R}^{20}$: five environmental features
per cell and month, including ERA5 wind and waves
\cite{hersbach2020era5} and ORAS5 ice thickness, concentration, and
temperature \cite{zuo2019oras5}; six geometric features for location and
goal relation; and nine static vessel features from a $414$-vessel
registry. We apply $\log(1+x)$ to ice variables, standardize the
remaining features using training-split statistics, and clip them at
$\pm5\sigma$.

\subsection{Experimental Setup}\label{sec:protocol}
\vspace{-2pt}
We use vessel-disjoint splits, ensuring that no vessel appears in more
than one of the training, validation, and test sets. This evaluates
generalization to unseen demonstrators. We additionally evaluate
temporal robustness using a temporal-shift split that holds out the
final months of trajectories from vessels observed during training.
The resulting splits contain
$1{,}939/246/833/168$ episodes and
$134/20/41/39$ vessels for train, validation, test, and temporal-shift,
respectively. All models are evaluated on the same held-out query decisions. For methods that infer latent context, three episodes per test vessel are reserved as support trajectories for estimating $z$ and are excluded from evaluation.

\subsection{Reward Models and Baselines}
\label{sec:methods-models}
\vspace{-2pt}
We use capacity-matched comparisons to separate nonlinear reward capacity
from per-vessel latent context. MCE-IRL versus AIRL tests nonlinear
capacity with a shared reward, whereas AIRL versus PEMIRL tests latent
context at fixed nonlinear architecture, optimizer, and training budget.

MCE-IRL learns a shared linear reward
$r_\theta(s)=\theta^\top\phi(s)$ by maximizing demonstration likelihood
under the soft-optimal policy in Eq.~\ref{eq:softvi}. AIRL learns a shared
nonlinear reward adversarially against a PPO policy. PEMIRL augments AIRL
with a recurrent inference network that infers a vessel-specific Gaussian
context $z\in\mathbb{R}^{8}$ from support trajectories and conditions
both $f(s,a,z)$ and $\pi(a\mid s,z)$ on $z$~\cite{MHAIRL}.

We additionally evaluate canonical AIRL \cite{fu2018learning}, GAIL
\cite{ho2016generative}, BC, sequence-policy baselines (LSTM and
Transformer), and the original PEMIRL implementation \cite{yu2019meta}.
All methods use the same data, splits, action masking, and evaluation
protocol.

\subsection{Learned Rewards Evaluation}
\label{sec:methods-eval}
\vspace{-2pt}

We evaluate learned rewards along three complementary dimensions.
Predictive accuracy is measured by the log-likelihood of held-out expert
actions, reported with both decision-level and vessel-level
aggregation. Behavioral fidelity measures whether the learned reward
reproduces realistic navigation trajectories on held-out
origin--destination pairs using trajectory similarity and task-completion
metrics. Reward transfer measures whether the learned reward generalizes
beyond the training policy by freezing the reward, training a masked
PPO agent from scratch, and evaluating its navigation performance. All
stochastic results are reported as mean $\pm$ standard deviation over
three random seeds.

\section{Results}
\label{sec:results}
\vspace{-2pt}
Our central finding is that behavioral heterogeneity alone does not imply
useful latent context. Once a shared reward has sufficient nonlinear
capacity to model observable route and environmental factors, adding
per-vessel context provides no benefit for held-out prediction, route
generation, or reward transfer.

\subsection{Nonlinear Reward Capacity \textit{Versus} Latent Context}
\label{sec:rq1}
\vspace{-2pt}

We first test whether PEMIRL's predictive gains arise from its
per-vessel latent context or from increased reward capacity. The standard
comparison between PEMIRL and MCE-IRL changes both factors simultaneously,
making the source of improvement ambiguous. We therefore introduce AIRL
as a capacity-matched control. AIRL uses the same nonlinear reward
architecture, optimizer, and training budget as PEMIRL, but removes the
per-vessel latent context and its associated inference objective.

Under the standard comparison, PEMIRL improves over MCE-IRL on unseen
vessels (Table~\ref{tab:landscape}). However, the capacity-matched
comparison reveals that the gain comes from nonlinear reward capacity
rather than latent context. AIRL improves over MCE-IRL by
$30.3\% \pm 4.1\%$ per decision and $50.9\% \pm 1.1\%$ per vessel without latent context, whereas adding PEMIRL's latent context reduces likelihood relative to AIRL. Feature-expectation error follows the same ordering, with AIRL achieving the lowest error.

These results show that PEMIRL's apparent advantage over MCE-IRL is
primarily explained by nonlinear reward capacity rather than by
vessel-specific latent context. Per-vessel comparisons across the $41$ test vessels show the same ordering,
with AIRL outperforming MCE-IRL on $36$ of $41$ vessels while PEMIRL does not consistently improve over AIRL. Temporal-shift, support-trajectory, and trajectory-length checks preserve
the same conclusion. The main benefit of AIRL is reducing extreme failures:
on matched test-split queries, the worst-performing MCE-IRL vessel scores
$-5.04$ nats per decision, while AIRL improves the same vessel to
$-1.51$, with no AIRL vessel below $-2.28$. AIRL improves over MCE-IRL on
$36/41$ test vessels. We further compare against imitation models using the
same held-out per-decision likelihood metric. Behavior cloning (BC) and
sequence models achieve higher predictive likelihood, with the Transformer
reaching $-0.93$ and the LSTM $-0.82$. This indicates that within-voyage
history captures predictive information more effectively than PEMIRL's
cross-voyage latent context. GAIL reaches $-1.38$, below BC under the same
protocol, consistent with the distinction between occupancy matching and
direct next-action prediction \cite{orsini2021matters}.

\subsection{Route Decoding and Reward Transfer}
\label{sec:rq2}
\vspace{-2pt}

\begin{table*}[t]
\centering
\caption{Comparison of reward and imitation models under a shared
evaluation protocol. LL denotes held-out per-decision log-likelihood.
Results are mean $\pm$ std over seeds $\{0,1,2\}$. Transfer requires a
frozen scalar training signal: BC and sequence policies provide only
policies, whereas GAIL transfers its discriminator-derived surrogate
$-\log(1-D(s,a))$. Route decoding is restricted to rewards evaluated
through the MCE-style soft-value-iteration decoder.}
\label{tab:landscape}
\setlength{\tabcolsep}{3pt}
\renewcommand{\arraystretch}{1.0}
\small
\begin{tabular}{llccccc}
\toprule
Model & Recovered reward & LL micro $\uparrow$ & LL macro $\uparrow$ &
Hausd. (km) $\downarrow$ & Reach $\uparrow$ & Transfer $\uparrow$ \\
\midrule
Masked-uniform chance & --- & $-1.90$ & --- & --- & --- & $\ms{0.09}{0.00}$ \\
\midrule
MCE-IRL & linear, shared & $\ms{-1.76}{0.01}$ & $\ms{-1.85}{0.06}$ &
  $\mathbf{45.60}$ & $0.00$ & $\ms{0.18}{0.01}$ \\
AIRL & nonlinear, shared & $\ms{-1.49}{0.02}$ & $\ms{-1.44}{0.06}$ &
  $\ms{231.30}{6.20}$ & $\ms{0.16}{0.05}$ & $\ms{0.59}{0.01}$ \\
PEMIRL & nonlinear, per-vessel & $\ms{-1.62}{0.04}$ & $\ms{-1.62}{0.06}$ &
  $\ms{250.90}{4.80}$ & $\ms{0.04}{0.02}$ & $\ms{0.59}{0.05}$ \\
Canonical AIRL & shaped, state-only $g$ & $\ms{-1.43}{0.01}$ & $\ms{-1.36}{0.02}$ &
  $\ms{233.30}{16.90}$ & $\ms{0.45}{0.02}$ & $\mathbf{\ms{0.81}{0.06}}$ \\
Original meta-IRL impl. & nonlinear, per-vessel & $\mathbf{\ms{-1.29}{0.03}}$ & $\mathbf{\ms{-1.14}{0.03}}$ &
  $\ms{176.60}{15.40}$ & $\mathbf{\ms{0.53}{0.02}}$ & $\ms{0.35}{0.24}$ \\
\midrule
GAIL & discriminator surrogate (not IRL reward) & $\ms{-1.38}{0.01}$ & $\ms{-1.30}{0.02}$ &
  --- & --- & $\ms{0.62}{0.11}$ \\
BC-MLP & none (imitation) & $\ms{-1.26}{0.01}$ & $\ms{-1.17}{0.02}$ &
  --- & --- & --- \\
seq-Transformer & none (imitation) & $\ms{-0.93}{0.00}$ & $\ms{-0.83}{0.01}$ &
  --- & --- & --- \\
seq-LSTM & none (imitation) & $\mathbf{\ms{-0.82}{0.01}}$ & $\mathbf{\ms{-0.77}{0.03}}$ &
  --- & --- & --- \\
\bottomrule
\end{tabular}
\renewcommand{\arraystretch}{1}
\end{table*}

\begin{figure}[t]
\centering
\includegraphics[width=\columnwidth, height=1.5in]{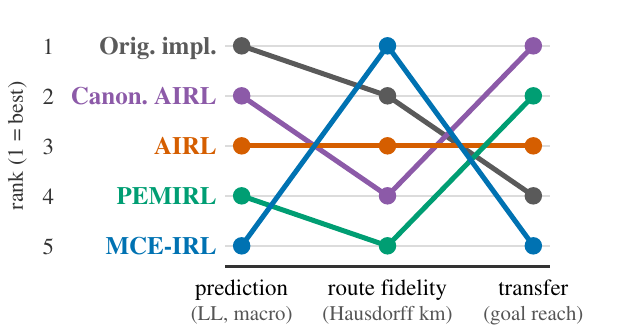}
\caption{Reward-derived evaluation criteria rank models differently
across held-out action likelihood, decoded-route fidelity, and frozen-signal
policy transfer. MCE-IRL produces the most realistic routes but performs
worst on prediction and transfer, while nonlinear reward formulations
reverse this trend. PEMIRL offers no improvement over context-free AIRL
on any criterion.}
\label{fig:criteria}
\end{figure}

We next evaluate whether each recovered reward generates realistic full
routes and can train a new policy from scratch. The model ranking changes
substantially across these criteria
(Table~\ref{tab:landscape}; Figure~\ref{fig:criteria}).

\begin{figure*}[t]
\centering
\includegraphics[width=\textwidth, height=2.4in]{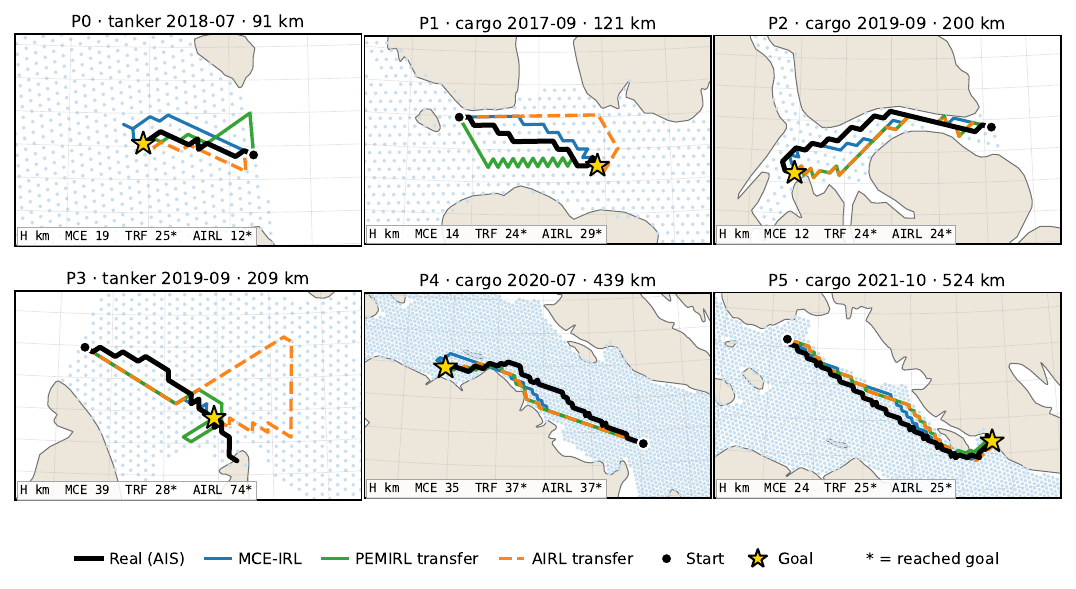}
\caption{Decoded routes for six held-out origin--destination pairs. Each panel compares the real AIS trajectory with the MCE-IRL-decoded route and with fresh PPO agents trained on the recovered rewards from AIRL and PEMIRL. Hausdorff distance to the real trajectory is reported for each route. MCE-IRL is shown for corridor comparison despite not reaching the goal.}
\label{fig:gallery}
\end{figure*}

Route decoding does not follow the likelihood ranking. MCE-IRL produces
the most realistic complete routes, with $45.6$\,km Hausdorff distance
and length ratio $0.77$, whereas AIRL and PEMIRL produce trajectories
$12$--$14$ times the correct length (Figure~\ref{fig:gallery}). Their nearly identical route metrics indicate that this failure stems
from the nonlinear reward formulation rather than latent context. The ordering also
persists when both rewards are scored with MCE-IRL's soft value iteration
($947$\,km for AIRL and $1{,}030$\,km for PEMIRL). Thus, improved
next-step prediction does not imply realistic long-horizon rollouts.

Reward transfer gives a different result. A newly initialized PPO agent
trained on a frozen nonlinear reward reaches held-out goals at steady
state on $59.1\% \pm 1.1\%$ of episodes for AIRL and
$59.2\% \pm 5.2\%$ for PEMIRL, compared with $8.9\%$ for a
uniform-random policy. The corresponding jointly trained PEMIRL
generator reaches only $4.2\% \pm 1.5\%$, showing that a recovered reward
can support a new policy even when the policy trained alongside it does
not. Figure~\ref{fig:gallery} shows both effects on six held-out
origin--destination pairs: MCE-IRL follows the correct corridor but never
terminates at the goal, whereas agents trained on the frozen AIRL and
PEMIRL rewards reach the goal along less direct routes and are visually
indistinguishable from each other.

GAIL provides an additional transfer control because its discriminator
defines a frozen surrogate, $-\log(1-D(s,a))$: PPO trained on this signal
reaches held-out goals at $61.7\% \pm 10.6\%$. BC and sequence models
provide only conditional policies, not a scalar reward to freeze for PPO;
we omit GAIL route decoding because our decoder requires an MCE-style
soft-value-iteration reward.

MCE-IRL transfers poorly under the same optimization procedure. Its
goal-reaching rate peaks at $0.385$ and falls to $0.175$ as PPO learns
to loiter in high-reward regions rather than reach the goal. This is
consistent with reward hacking under optimization pressure
\cite{amodei2016concrete,skalse2022defining}.

We finally test whether this conclusion depends on implementation.
Canonical AIRL (Eq.~\ref{eq:airl}) predicts better than AIRL and PEMIRL
($-1.434 \pm 0.012$) and transfers best
($0.809 \pm 0.061$ for its state-only reward $g$). Conversely, the
original meta-IRL implementation gives the strongest likelihood
($-1.288 \pm 0.026$) and highest decoded goal reach
($0.530 \pm 0.017$), but its frozen reward transfers unreliably
($0.350 \pm 0.240$) across seeds. Reliable reward transfer depends more on parameterization and optimization stability than on latent-context modeling.

\subsection{Sources of Behavioral Heterogeneity}
\label{sec:rq3}
\vspace{-2pt}

We next ask why PEMIRL does not benefit from per-vessel context despite
substantial vessel-level heterogeneity. Vessel identity explains
$\eta^2 = 0.33$ of the variance in behavioral descriptors
\cite{lakens2013calculating}, far more than any category label
($\eta^2 = 0.085$ for size class), and $\eta^2 = 0.34$ after excluding
short-haul voyages. However, this heterogeneity is largely explained by
observable route and environmental factors already available to the shared
reward.

Across $123$ vessels with at least five voyages, only speed is strongly
heterogeneous by vessel ($F = 2.47$); route-shape descriptors remain
homogeneous ($F \leq 0.38$). Vessels therefore follow similar paths but
move at different speeds. Grouping the same voyages by
origin--destination route produces more variation than grouping them by
vessel ($F = 2.45$ versus $0.83$). This pattern mirrors the variable-centered \textit{versus} person-centered distinction~\cite{laursen2006, howard2018}: PEMIRL's per-vessel context is a person-centered representation, while AIRL's shared nonlinear reward is variable-centered. Here, the person-centered grouping is the weaker one; the apparent vessel effect is primarily a route-assignment effect. On the route-matched subset ($82$ routes with at least three voyages, $n = 330$), removing each route's mean drops the vessel-level $F$-ratio from $1.03$ to $0.23$ overall and from $3.66$ to $0.44$ for speed, indicating that the apparent vessel effect primarily reflects route-assignment.

Sea ice provides an observable explanation for this pattern. The same corridor is traversed quickly in open water and slowly under heavy ice. Consistent with this account, an ice-free Gulf of St.~Lawrence fleet shows no heterogeneous grouping with $F > 1$. Route
and ice are included in $\phi(s)$, so a nonlinear shared reward can
represent their effects without identifying the vessel. This explains why AIRL benefits from additional reward capacity while PEMIRL gains little from latent context.

The null performance effect does not arise because the latent context
failed to train. Across $41$ test vessels, the standard deviation of the posterior mean ranges from $0.028$ to $0.075$ over latent dimensions, and changing $z$ alone
changes the reward by $60\%$ as much as changing the full state--action
input. A linear probe predicts the nine observable vessel-static
features from $z$ with $R^2 = 0.63$. The learned context is therefore
active, but it primarily re-encodes information already available to the
shared reward.

We next test whether this redundancy is caused by vessel metadata in the
observation. Removing all nine static vessel features leaves an
$11$-dimensional $\phi(s)$, creating a setting favorable to
latent-context. Yet AIRL remains stronger than PEMIRL
($-1.383$ versus $-1.566$ macro likelihood), while AIRL itself does not
degrade relative to the full-state setting ($-1.441$). With metadata
hidden, vessel identity is weakly recoverable from behavior alone
($R^2 = 0.14$).

Finally, the ice-free Gulf of St.~Lawrence replication follows the same
prediction. The capacity gain shrinks to $+8.5\%$, compared with
$+30.3\%$ in the Arctic, while the latent-context step remains negative
($-12.8\%$). Across both regions, context fails when observable factors
account for the relevant behavioral variation. These results suggest that demonstrator heterogeneity warrants latent context only when it remains unexplained after conditioning on observable state, route, and environment.

\subsection{A Practical Context-Need Diagnostic}
\label{sec:rq4}
\vspace{-2pt}

The preceding analyses suggest a diagnostic for deciding whether latent
context is likely to add information beyond the observed state. This
diagnostic first measures demonstrator-level heterogeneity, identifies the
descriptors in which it occurs, and recomputes it after conditioning on
observed covariates. In our setting, this isolates speed as the primary
heterogeneous descriptor ($F = 2.47$ versus $F \leq 0.38$ for route shape)
and shows that conditioning on route substantially reduces this variation
($F: 1.03 \rightarrow 0.23$ overall, $3.66 \rightarrow 0.44$ for speed).
Persistent residual variation would motivate a latent-context model,
whereas variation that largely disappears favors a capacity-matched shared
reward as the primary baseline.

When a latent-context model is trained, the learned context should also be
audited rather than treated as evidence of useful hidden structure by
itself. Posterior variation and reward sensitivity establish whether the
context influences the model, while decoding observed covariates tests
whether it captures additional information or re-encodes the observed
state. In our experiments, the learned context influences the reward yet
largely reflects observable vessel properties, and the hidden-covariate
ablation confirms that this decoding is not solely due to information
already exposed in the state. Together, these analyses distinguish
genuinely hidden behavioral structure from latent representations that
primarily capture observable features.

\section{Discussion}
\label{sec:discussion}
\vspace{-2pt}
The central distinction is between heterogeneity explained by the observed state
and heterogeneity that remains unobserved. Demonstrators may differ
substantially while those differences are attributable to route,
environment, or other variables already available to a shared reward. In
such settings, a latent-context formulation need not improve reward
learning, even when its inferred context is active. Latent context is
most compelling when it captures decision-relevant variation that the
state does not adequately represent. This is a property of both the
dataset and the state representation. The same trajectories could support a different modeling choice if relevant environmental variables were omitted or if unobserved factors, such as operator intent, influenced behavior. Similar considerations arise in other real-world settings, such as autonomous driving, where observed differences may reflect road and traffic conditions rather than unobserved driver preferences.

\paragraph{Implications for Evaluation.}
Latent-context IRL should be evaluated against a capacity-matched shared
control. Comparing a context-conditioned model only with a simpler shared
baseline can confound the value of nonlinear reward capacity with the
value of latent context. Evaluation should also extend beyond held-out
next-action likelihood. Predictive accuracy, long-horizon route decoding,
and reward transfer measure different properties of a learned system and
may favor different models. Evaluating a frozen reward under a newly
trained policy provides a direct test of how it behaves under downstream
optimization.

\paragraph{Scope and Future Directions.}
Our conclusions concern per-vessel context in a direction-only route-choice
MDP. Speed accounts for much of the observed vessel-level variation but
is not an action in this formulation, so the results do not establish
whether latent information could improve operational speed control or
joint route--speed planning. Other task definitions may also be more
appropriate when variation is indexed by route, cargo, company, season,
or time rather than vessel identity. Likewise, partially observed factors
such as operator intent may require history-conditioned policies or belief
states instead of a static per-vessel context. Hierarchical reward models,
constraint-based formulations, and models that explicitly represent
dynamics are complementary ways to model temporal subgoals, operational
constraints, or environment mismatch.

\paragraph{Implications for Arctic Navigation.}
For AI-assisted Arctic route analysis and decision support, our results
favor adaptation through observed, decision-relevant conditions over
unvalidated per-vessel latent representations. Known operational
requirements should be encoded as explicit constraints, enabling
transparent and auditable recommendations. Per-vessel latent context
should be introduced only when diagnostic evidence shows that observable
features cannot explain behavioral variation. This provides a principled
basis for developing accountable AI decision-support systems in
safety-critical Arctic operations.

\paragraph{Limitations.}
This study uses cargo and tanker voyages in a single graph
MDP, so the findings may not generalize to other environments, feature
representations, or reward formulations. Reward-transfer results are
specific to the MDP and PPO configuration. Finally, observed
trajectories reflect behavior rather than underlying operational intent,
so cross-domain evaluation is needed to determine when latent context
provides benefits beyond observable state information.

\section{Conclusion}
\label{sec:conclusion}
\vspace{-2pt}
This work shows that observed behavioral differences alone do not justify the introduction of latent-context inverse reinforcement learning. Through a controlled evaluation in Arctic shipping, we separate the benefits of expressive reward modeling from those of per-vessel latent context and show how an observability analysis can help assess whether personalization is empirically warranted. For AI-assisted navigation, this highlights the importance of grounding adaptive systems in the information available from the operating environment before attributing behavior to hidden factors. Latent representations should be introduced only when they provide additional decision-relevant information beyond observable conditions. More broadly, this work supports the evidence-based use of latent-context models, in which additional model complexity should be justified by information not already represented in the observed state.

\vspace{.3cm}\noindent
\hfill{\Large \textit{\textbf{Acknowledgments}}}\hfill
\vspace{.1cm}\\
This work was supported by the Natural Sciences and Engineering Research Council of Canada --- NSERC, the Faculty of Computer Science of Dalhousie University, and the Conselho Nacional de Desenvolvimento Cient\'{i}fico e Tecnol\'{o}gico --- CNPq, Brasil.

\vspace{.3cm}\noindent
\hfill{\Large \textit{\textbf{Responsible AI Use}}}\hfill
\vspace{.1cm}\\
Generative AI technologies supported initial manuscript drafting and the creation of preliminary code scaffolding. The authors independently reviewed, verified, and revised all AI-assisted material and retain full responsibility for the study's methods, claims, analyses, interpretations, figures, and reported results.

\putbib
\end{bibunit}

\appendix
\begin{bibunit}
\vspace{.3cm}\noindent
\hfill{\Large \textit{\textbf{Appendix}}}\hfill
\vspace{.1cm}
\section{A. Reproducibility Details for Data and Evaluation}
\label{app:protocol}

The main paper specifies the trajectory corpus, masked H3 graph MDP,
state representation, and vessel-disjoint split design. This section
records implementation details and sample counts needed to reproduce
the reported evaluations.

\paragraph{Graph and action masking.}
The Arctic navigation graph spans $60.0^\circ$--$77.0^\circ$\,N and
$127.9^\circ$--$64.5^\circ$\,W and contains $14{,}206$ navigable H3
resolution-6 cells and $39{,}053$ undirected adjacency edges. The
action space has seven actions: action $0$ remains in the current cell,
and actions $1$--$6$ select neighboring cells in clockwise bearing
order from north. At coastal and pentagon-boundary cells, unavailable
actions are masked by adding $-10^{9}$ to their logits. Thus, masked
actions have zero probability in every policy, likelihood calculation,
chance baseline, and decoded route. We recover actions from consecutive
discretized cells and use $H=512$ and $\gamma=0.99$ throughout.

\paragraph{Preprocessing and features.}
The corpus comprises cargo and tanker AIS trajectories from July--October
of 2016--2024. After voyage segmentation, H3 discretization, removal of
repeated cells, bridging of single-cell gaps, and exclusion of voyages
with fewer than five transitions, it contains $3{,}186$ demonstrations
($2{,}375$ cargo; $811$ tanker). Each state representation has 20
components: five monthly ERA5/ORAS5 environmental variables, six
location- and goal-relative geometric terms, and nine vessel-registry
attributes. Sea-ice variables use $\log(1+x)$; all other features are
standardized using training-split statistics and clipped to
$\pm 5\sigma$. Missing registry attributes, affecting approximately
$7\%$ of demonstration vessels, are set to zero.

\paragraph{Evaluation populations.}
Training, validation, and test splits are vessel-disjoint. The temporal
shift uses later months from vessels represented in training. For
context-conditioned models, three episodes per evaluation vessel are
reserved as support trajectories and are never included in likelihood
scoring. All model comparisons use the matched query population: only
decisions scored by every model under the relevant protocol are retained.

\begin{table}[H]
\centering
\caption{Dataset splits and matched query populations. The train,
validation, and test splits are vessel-disjoint and together cover
$3{,}018$ voyages from $195$ vessels; the temporal-shift split holds the
remaining $168$ voyages of the $3{,}186$-voyage corpus.}
\label{tab:dataset}
\setlength{\tabcolsep}{4pt}
\renewcommand{\arraystretch}{1.1}
\small
\begin{tabular}{lrrrr}
\toprule
Split & Episodes & Vessels & Decisions & Query dec. \\
\midrule
Train          & $1{,}939$ & $134$ & $151{,}782$ & --- \\
Validation     & $246$     & $20$  & $14{,}894$  & --- \\
Test           & $833$     & $41$  & $70{,}995$  & $62{,}127$ \\
Temporal shift & $168$     & $39^{\dagger}$ & $17{,}411$ & $8{,}508$ \\
\bottomrule
\multicolumn{5}{l}{\footnotesize $^{\dagger}$32 are training vessels
evaluated in unseen months;}\\
\multicolumn{5}{l}{\footnotesize \phantom{$^{\dagger}$}the other 7 occur in
no other split, giving 202 corpus-wide.}
\end{tabular}
\renewcommand{\arraystretch}{1}
\end{table}

\section{B. Training and Optimization Details}
\label{app:hparams}

This section gives the settings needed to reproduce model training.
All models use the same MDP, action masks, state features,
vessel-disjoint splits, matched query sets, and random seeds
$\{0,1,2\}$. AIRL and PEMIRL are trained on the same 1,000
demonstrations. Table~\ref{tab:hparams} lists the main settings.

\paragraph{MCE-IRL training.}
We train MCE-IRL with Adam for 200 iterations. We select the model
checkpoint with the best validation log-likelihood. This checkpoint
occurs at iteration 40 and has validation log-likelihood $-1.754$.
The learned parameter vectors are highly consistent across seeds, with
pairwise cosine similarity above $0.9999$.

For each demonstration group, soft value iteration uses the
corresponding goal, year, month, vessel, horizon, and environmental
features. Thus, the policy is evaluated under the conditions of the
voyage being modeled.

\paragraph{AIRL and PEMIRL training.}
AIRL and PEMIRL are trained for 500 outer iterations without early
stopping. Each iteration generates 64 rollouts. PEMIRL samples 16
latent contexts and generates four rollouts for each context.

For reward-transfer experiments, we freeze the learned reward and
train a new PPO agent from scratch. PPO uses 200 iterations with 64
rollouts per iteration. We compute all reported evaluation metrics on
CPU.

\paragraph{PEMIRL stability settings.}
The original PEMIRL information-maximization objective was unstable in
our graph-navigation setting. In particular, the objective uses the
sum of rewards along a trajectory, and this sum can grow without bound.
We therefore use the following stability settings in all default PEMIRL
experiments:
\begin{enumerate}
\item Clamp the trajectory return used by the information objective to
      the range $[-10,10]$.
\item Standardize the information-objective advantage within each
      minibatch.
\item Clip discriminator and posterior-network gradient norms to $1.0$.
\item Use the discriminator logit $f$ directly as the PPO reward,
      rather than using $\sigma(f)$.
\item Set the information-maximization weight to $\eta=0.01$.
\item Start posterior-network updates after the first 10 outer
      iterations.
\end{enumerate}

For the information-maximization sensitivity analysis, we change only
the weight to $\eta=0.1$ and retain the other five settings.

\begin{table}[H]
\centering
\caption{Main training and architecture settings. GAE denotes
generalized advantage estimation. The latent clamp bounds sampled
PEMIRL context values.}
\label{tab:hparams}
\setlength{\tabcolsep}{2pt}
\renewcommand{\arraystretch}{1.1}
\small
\begin{tabular}{lll}
\toprule
Component & Setting & Value \\
\midrule
MCE-IRL
  & Learning rate & $0.05$ \\
  & $\ell_2$ penalty & $10^{-4}$ \\
  & Training iterations & $200$ \\
\midrule
AIRL
  & Discriminator arch. & MLP $(256,256)$ \\
  & Generator arch. & MLP $(256,256)$ \\
  & Gradient-penalty weight & $10$ \\
\midrule
PEMIRL
  & Latent dimension & $8$ \\
  & Latent clamp & $\pm2$ \\
  & Posterior arch. & Bi-LSTM $(128)$ \\
  & Information weight $\eta$ & $0.01$ \\
  & Return clamp & $10$ \\
  & Posterior warm-up & $10$ iterations \\
\midrule
Adversarial
  & Outer iterations & $500$ \\
  & Rollouts per iteration & $64$ \\
  & Training demonstrations & $1{,}000$ \\
  & Random seeds & $\{0,1,2\}$ \\
\midrule
PPO
  & Clip ratio & $0.2$ \\
  & GAE $\lambda$ & $0.95$ \\
  & Optimization epochs & $8$ \\
  & Entropy coefficient & $0.01$ \\
  & Value coefficient & $0.5$ \\
  & Learning rate & $3\times10^{-4}$ \\
\midrule
Reward transfer
  & Support episodes per vessel & $3$ \\
  & PPO iterations & $200$ \\
  & Rollouts per iteration & $64$ \\
\bottomrule
\end{tabular}
\renewcommand{\arraystretch}{1}
\end{table}

\section{C. Sensitivity Analysis of the Main Comparison}
\label{app:robustness}

This section tests whether the main comparison changes under different
evaluation splits, aggregation methods, support-trajectory choices, or
trajectory-length filtering. The main paper reports the central
comparison; here we provide the full results.

\paragraph{Evaluation splits and aggregation.}
Table~\ref{tab:ladder} reports held-out action log-likelihood on the
standard test split and the temporal-shift split. We report two
averages. Micro aggregation gives equal weight to every decision.
Macro aggregation gives equal weight to every vessel. We also report
test feature-expectation error (FEE), where lower values are better.

The same model ordering appears in each split and aggregation:
AIRL has the best log-likelihood, followed by PEMIRL and MCE-IRL.
The temporal-shift results therefore do not change the main conclusion.

\paragraph{Support and horizon checks.}
Table~\ref{tab:variants} tests two possible sources of evaluation
bias. First, it compares randomly selected support voyages with causal
support voyages, defined as each vessel's earliest available voyages.
Second, it excludes 37 episodes that reach the horizon limit of 512
steps.

These checks leave the model ordering unchanged. In particular, causal
support does not improve PEMIRL enough to close its gap with AIRL.
This suggests that the central result is not explained by information
from later support voyages or by horizon-truncated episodes.

\paragraph{Per-vessel performance.}
We also evaluate each test vessel separately. On the matched test set,
the lowest MCE-IRL score is $-5.04$ nats per decision. AIRL obtains
$-1.51$ nats on that same vessel, and its lowest score across all test
vessels is $-2.28$ nats per decision. Complete per-vessel and
per-episode results are available in
\texttt{runs/eval/per\_episode\_ll*.json}.

\begin{table}[H]
\centering
\caption{Held-out per-decision log-likelihood by split and aggregation
(higher is better), reported as mean $\pm$ standard deviation across
three seeds. The empirical masked-action chance log-likelihood is
$-1.90$ on the test split and $-1.91$ on the temporal-shift split.
FEE denotes test feature-expectation error (lower is better).}
\label{tab:ladder}
\setlength{\tabcolsep}{1.5pt}
\renewcommand{\arraystretch}{1.15}
\small
\begin{tabular}{llccc}
\toprule
Split & Average & MCE-IRL & AIRL & PEMIRL \\
 & & linear & nonlinear & $+$ context \\
\midrule
\multirow{2}{*}{Test}
  & micro & $\ms{-1.76}{0.01}$ &
            $\mathbf{\ms{-1.49}{0.02}}$ &
            $\ms{-1.62}{0.04}$ \\
  & macro & $\ms{-1.85}{0.06}$ &
            $\mathbf{\ms{-1.44}{0.06}}$ &
            $\ms{-1.62}{0.06}$ \\
\midrule
\multirow{2}{*}{Temporal shift}
  & micro & $\ms{-1.72}{0.04}$ &
            $\mathbf{\ms{-1.47}{0.03}}$ &
            $\ms{-1.59}{0.04}$ \\
  & macro & $\ms{-1.85}{0.08}$ &
            $\mathbf{\ms{-1.42}{0.05}}$ &
            $\ms{-1.60}{0.06}$ \\
\midrule
\multicolumn{2}{l}{Test FEE $\downarrow$}
  & $826$ & $\mathbf{\ms{617}{35}}$ & $\ms{677}{27}$ \\
\bottomrule
\end{tabular}
\renewcommand{\arraystretch}{1}
\end{table}

\begin{table}[H]
\centering
\caption{Test log-likelihood under support and trajectory-length
variants, averaged over three seeds. R uses randomly selected support
voyages; C uses causal support, namely the earliest available voyages.
$-$T excludes the 37 horizon-truncated test episodes. The largest
standard deviation across seeds is $0.062$.}
\label{tab:variants}
\setlength{\tabcolsep}{4pt}
\renewcommand{\arraystretch}{1.15}
\small
\begin{tabular}{llcccc}
\toprule
Average & Model & R & R$-$T & C & C$-$T \\
\midrule
\multirow{3}{*}{Macro}
 & MCE-IRL & $-1.852$ & $-1.825$ & $-1.849$ & $-1.834$ \\
 & AIRL    & $\mathbf{-1.441}$ & $\mathbf{-1.433}$ &
              $\mathbf{-1.461}$ & $\mathbf{-1.457}$ \\
 & PEMIRL  & $-1.621$ & $-1.619$ & $-1.590$ & $-1.588$ \\
\midrule
\multirow{3}{*}{Micro}
 & MCE-IRL & $-1.757$ & $-1.675$ & $-1.752$ & $-1.674$ \\
 & AIRL    & $\mathbf{-1.492}$ & $\mathbf{-1.475}$ &
              $\mathbf{-1.496}$ & $\mathbf{-1.482}$ \\
 & PEMIRL  & $-1.620$ & $-1.614$ & $-1.620$ & $-1.614$ \\
\bottomrule
\end{tabular}
\renewcommand{\arraystretch}{1}
\end{table}

\section{D. Feature-Hiding Ablation}
\label{app:hidden}

This ablation tests whether latent context becomes useful when static
vessel information is removed from the observed state. The full state
contains 20 features, including nine vessel attributes. In the hidden
setting, we remove all nine vessel attributes and retain 11 features:
five environmental variables, normalized latitude, $(\cos,\sin)$
longitude, distance to the goal, and $(\cos,\sin)$ bearing to the goal.

We adjust network input sizes to match the reduced feature vector. The
context-conditioned discriminator receives 26 inputs, and the
context-free discriminator receives 18 inputs. All other settings are
unchanged: the dataset splits, feature standardization procedure,
training budget, and random seeds are the same as in the full-feature
experiments.

Before running this experiment, we specified the hypothesis that latent
context might compensate for omitted vessel metadata. This is the only
preregistered analysis in the study. The heterogeneity analysis in
Section~F was developed after the main experiments.

\paragraph{Results.}
Table~\ref{tab:hidden} compares held-out test log-likelihood using the
full and hidden state representations. Removing vessel attributes does
not improve PEMIRL relative to AIRL. AIRL remains the best-performing
model under both micro and macro aggregation.

\begin{table}[H]
\centering
\caption{Feature-hiding results on the test split. Values are mean
per-decision log-likelihood across three seeds; higher is better. The
largest seed standard deviation is $0.063$.}
\label{tab:hidden}
\setlength{\tabcolsep}{4pt}
\renewcommand{\arraystretch}{1.15}
\small
\begin{tabular}{lcccc}
\toprule
 & \multicolumn{2}{c}{Full ($D=20$)} &
   \multicolumn{2}{c}{Vessel feat.\ hidden ($D=11$)} \\
\cmidrule(lr){2-3}\cmidrule(lr){4-5}
Model & Micro & Macro & Micro & Macro \\
\midrule
MCE-IRL & $-1.757$ & $-1.852$ & $-1.896$ & $-2.083$ \\
AIRL    & $-1.492$ & $-1.441$ &
          $\mathbf{-1.447}$ & $\mathbf{-1.383}$ \\
PEMIRL  & $-1.620$ & $-1.621$ & $-1.573$ & $-1.566$ \\
\bottomrule
\end{tabular}
\renewcommand{\arraystretch}{1}
\end{table}

\paragraph{Why MCE-IRL cannot use these features.}
The removed vessel attributes are constant within a voyage. Therefore,
a linear MCE-IRL reward can assign them only a constant reward offset
within that voyage. This offset changes values but not action
probabilities.

\noindent\textbf{Proposition.}
Let $\phi(s) = [\psi(s);m]$, where $m$ is constant for every state in
an episode, and let
$r_\theta(s) = \theta^\top\phi(s)$. The soft value iteration policy for
that episode is independent of $m$ and its corresponding parameter
vector $\theta_m$.

\noindent\textbf{Proof.}
Write
$r_\theta(s) = \theta_\psi^\top\psi(s) + c$,
where $c = \theta_m^\top m$ is constant within the episode. At time
$t$, adding $c$ to every state reward adds the same quantity,
$c(H-t)$, to all action-values $Q_t(s,a)$ and to the state value
$V_t(s)$. The policy depends on their difference:
\[
\log \pi_t(a\mid s) = Q_t(s,a) - V_t(s).
\]
The shared offset therefore cancels. The same argument applies with
the goal boundary condition $V_t(g)=0$. Hence, episode-constant
metadata cannot change the MCE-IRL action policy.
\hfill$\square$

The MCE-IRL result should therefore not be interpreted as evidence
that the removed vessel attributes directly supported route choice in
the linear model. More importantly, hiding those attributes does not
give PEMIRL an advantage over the capacity-matched AIRL baseline.

\section{E. Context Audit}
\label{app:probes}

Context probes are evaluated on the seed-0 PEMIRL checkpoint across the 41 test vessels using posterior mean $\bar z_v$ inferred from three support episodes (Table~\ref{tab:probes}). Posterior spread measures the standard deviation of $\bar z_v$ across vessels. Reward sensitivity computes ratio $\sigma_z / \sigma_{sa}$ of reward standard deviation when varying $z$ versus varying state-action pairs $(s,a)$. Decoding probes fit linear regression models from $\bar z_v$ to the nine static vessel features, reporting variance-weighted mean $R^2$ values under full and metadata-hidden feature sets.

The measured posterior variation and feature decodability confirm that the context pathway is active and encodes observable vessel attributes. However, these metrics do not establish that the learned context provides useful hidden decision information for policy modeling.

\begin{table}[H]
\centering
\caption{Context probes evaluated on test vessels. $\sigma_z$ measures reward variation over posterior means $\bar z_v$ with $(s,a)$ fixed; $\sigma_{sa}$ measures reward variation over state-action pairs with $z$ fixed. Decoding probes report variance-weighted mean $R^2$ from linear regression of vessel statics on $\bar z_v$.}
\label{tab:probes}
\setlength{\tabcolsep}{3pt}
\renewcommand{\arraystretch}{1.15}
\small
\begin{tabular}{@{}l>{\raggedright\arraybackslash}p{2.6cm}c@{}}
\toprule
Probe & Quantity & Value \\
\midrule
Posterior spread & std of $\bar z_v$ across vessels, per dim.
  & $0.028$--$0.075$ \\
 & overall & $0.066$ \\
Reward sensitivity & $\sigma_z / \sigma_{sa}$ & $0.599$ \\
Decoding, full state & mean $R^2$, statics from $\bar z_v$ & $0.63$ \\
Decoding, metadata hidden & mean $R^2$ (per dim.\ $0.11$--$0.25$)
  & $0.14$ \\
\bottomrule
\end{tabular}
\renewcommand{\arraystretch}{1}
\end{table}

\section{F. Heterogeneity Methods}
\label{app:descriptors}

Per-voyage descriptors are calculated as follows:
\begin{itemize}
\item Straightness: Great-circle origin to destination displacement divided by trajectory path length.
\item Turn rate: Mean absolute heading change per step.
\item Mean speed and speed standard deviation: Derived from consecutive cell timestamps.
\item Loiter fraction: Share of trajectory steps with speed below $0.5$\,knots.
\end{itemize}

Effect sizes ($\eta^2$) are computed per grouping factor over $3{,}018$ voyages from $195$ vessels (Table~\ref{tab:eta}). Bias-adjusted $\varepsilon^2$ estimates \cite{olejnik2003generalized} preserve identical rank orderings. Variance ratios $F = \mathrm{Var}_{\mathrm{between}} / \mathrm{Var}_{\mathrm{within}}$ are calculated on z-scored descriptors. Inclusion thresholds require at least 5 voyages per group for vessel and route analyses, and at least 3 voyages for route-controlled analysis ($82$ routes, $n = 330$ voyages). The route-controlled protocol subtracts each route's mean descriptor value prior to grouping by vessel (Table~\ref{tab:decomp}).

\begin{table}[t]
\centering
\caption{Variance in per-voyage descriptors explained by grouping factors ($\eta^2$), computed over $3{,}018$ voyages from $195$ vessels.}
\label{tab:eta}
\setlength{\tabcolsep}{6pt}
\renewcommand{\arraystretch}{1.1}
\small
\begin{tabular}{lcc}
\toprule
Grouping factor & All voyages & Short-haul excl. \\
\midrule
Vessel identity & $\mathbf{0.330}$ & $\mathbf{0.341}$ \\
Size class      & $0.085$ & $0.084$ \\
Subtype         & $0.019$ & $0.020$ \\
Cargo vs.\ tanker & $0.007$ & $0.006$ \\
\bottomrule
\end{tabular}
\renewcommand{\arraystretch}{1}
\end{table}

\begin{table}[b]
\centering
\caption{Decomposition of vessel variance. Left: per-descriptor $F$-ratios by vessel. Right: variance ratios for route versus vessel groupings, including route-controlled vessel effects ($F > 1$ indicates that between-group variance exceeds within-group variance).}
\label{tab:decomp}
\setlength{\tabcolsep}{5pt}
\renewcommand{\arraystretch}{1.1}
\small
\begin{tabular}{lc@{\hspace{0.4em}}lc}
\toprule
Descriptor & $F$ by vessel & Grouping & $F$ \\
\midrule
straightness   & $0.14$ & by vessel & $0.83$ \\
turn rate      & $0.38$ & by O--D route & $\mathbf{2.45}$ \\
\textbf{mean speed} & $\mathbf{2.47}$ & vessel, route-ctrl.: & \\
speed std      & $0.97$ & \quad all descriptors & $1.03 \to \mathbf{0.23}$ \\
loiter fraction & $0.17$ & \quad mean speed & $3.66 \to \mathbf{0.44}$ \\
\bottomrule
\end{tabular}
\renewcommand{\arraystretch}{1}
\end{table}

The secondary evaluation region is the Gulf of St.\ Lawrence ($44.9^\circ$--$52.3^\circ$\,N, $71.7^\circ$--$54.7^\circ$\,W), discretized at H3 resolution 6 across all twelve months of 2024. Heterogeneity analysis covers $24{,}572$ voyages across nine vessel types. Model replication uses a goal-conditioned subset ($3{,}223$ voyages, $603$ vessels, $7{,}690$ states, $20{,}484$ edges) under identical split logic. Because environmental reanalysis data are unavailable for this region, feature vectors $\phi(s)$ include geometry, goal-relative terms, and vessel attributes only. Absolute likelihoods are not directly comparable to Arctic values, and only within-region model orderings are evaluated.

Table~\ref{tab:gsl} reports the replication. The within-region ordering
matches the Arctic: AIRL is best, MCE-IRL second, and PEMIRL last. We
express each step as the relative change in per-decision likelihood,
$\exp(\Delta\,\mathrm{LL})-1$. The capacity step from MCE-IRL to AIRL is
$+8.5\%$ on test and $+3.7\%$ under temporal shift, compared with
$+30.3\%$ in the Arctic. The latent-context step from AIRL to PEMIRL is
negative in both splits, $-12.8\%$ and $-12.5\%$. The reduced capacity
gain in an ice-free region, alongside an unchanged negative context step,
is the pattern predicted by the heterogeneity analysis. These runs use a
single seed and a reduced feature set, so they support the ordering
claim only, not the Arctic effect magnitudes.

\begin{table}[t]
\centering
\caption{Gulf of St.\ Lawrence replication (single seed, reduced feature
set). LL is held-out per-decision log-likelihood, higher is better; FEE
is feature-expectation error, lower is better. Absolute values are not
comparable to the Arctic results in Table~\ref{tab:ladder}.}
\label{tab:gsl}
\setlength{\tabcolsep}{5pt}
\renewcommand{\arraystretch}{1.15}
\small
\begin{tabular}{lccc}
\toprule
Model & Test LL & Temp.\ shift LL & Test FEE $\downarrow$ \\
\midrule
MCE-IRL & $-1.366$ & $-1.292$ & $863$ \\
AIRL    & $\mathbf{-1.285}$ & $\mathbf{-1.256}$ & $\mathbf{490}$ \\
PEMIRL  & $-1.422$ & $-1.390$ & $534$ \\
\bottomrule
\end{tabular}
\renewcommand{\arraystretch}{1}
\end{table}

\section{G. Reward Transfer Protocol}
\label{app:transfer}

The learned reward function is frozen, and a newly initialized action-masked PPO agent is trained for $200$ iterations of $64$ rollouts without access to expert demonstrations. For context-conditioned rewards, latent vector $z$ is inferred once per vessel from three support episodes and held constant. Performance is measured as the fraction of held-out test episodes reaching the goal cell within $H = 512$ steps. Steady-state performance is defined as the mean goal-reaching rate over the final $20\%$ of training iterations (40 iterations). A uniform-random policy over valid actions serves as baseline, achieving a $0.089$ goal-reaching rate.

\begin{figure}[H]
\centering
\includegraphics[width=\linewidth]{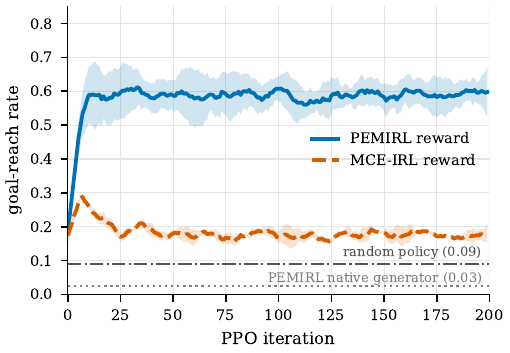}
\caption{Reward transfer training curves, mean $\pm$ standard deviation over three seeds. Baseline reference lines show the uniform-random valid policy ($0.089$) and the generator policy trained jointly with the PEMIRL reward ($0.025$, seed $0$).}
\label{fig:transfer}
\end{figure}

Figure~\ref{fig:transfer} illustrates transfer training dynamics. The linear reward exhibits non-monotonic transfer performance: goal-reaching efficiency peaks at $0.385$ before declining to a steady-state value of $0.175$. This drop occurs because the agent learns that remaining in high-reward states yields higher return than terminating at the goal. Nonlinear reward functions maintain stable goal-reaching performance throughout training.

A potential-shaping diagnostic, in which the transferred signal is
$r(s,a,s') + \gamma V(s') - V(s)$ rather than the raw reward, is reported
for the original implementation in Section~I.

\section{H. Reproducibility and Artifact Map}
\label{app:repro}

The code and data supplement contains no raw or processed AIS positional data, tracks, kinematics, or vessel names. Vessel identifiers are replaced with consistent opaque pseudonyms across all files. Artifacts marked with $\dagger$ in Table~\ref{tab:artifacts} are included in the supplement and can be inspected or re-derived without any AIS data. The remaining rows depend on the third-party licensed AIS corpus or on auxiliary experiment code. For those, the supplement ships the pipeline scripts and configuration files that document the exact procedure; the auxiliary experiment code is part of the full repository release.

\begin{table}[H]
\centering
\caption{Mapping of experimental results to repository artifacts. Artifacts marked with $\dagger$ are included in the code and data supplement and can be evaluated directly.}
\label{tab:artifacts}
\setlength{\tabcolsep}{3pt}
\renewcommand{\arraystretch}{1.15}
\small
\begin{tabular}{@{}l>{\raggedright\arraybackslash}p{4.1cm}@{}}
\toprule
Result & Artifact \\
\midrule
LL micro/macro (Tab.~\ref{tab:ladder}) &
  $\dagger$\,\texttt{runs/eval/}
  \texttt{per\_episode\_} \texttt{ll*.json},
  \texttt{robustness*} \texttt{.json} \\
Per-vessel dispersion &
  $\dagger$\,\texttt{runs/eval/}
  \texttt{per\_episode\_} \texttt{ll.json},
  \texttt{per\_episode\_} \texttt{ll\_noctx.json} \\
Paired per-vessel tests &
  $\dagger$\,\texttt{runs/eval/} \texttt{significance.json} \\
Support variants (Tab.~\ref{tab:variants}) &
  $\dagger$\,\texttt{runs/eval/} \texttt{support\_sweep*.json} \\
Route fidelity &
  $\dagger$\,\texttt{results/reference/} \texttt{routes\_*.csv} \\
Reward transfer (Fig.~\ref{fig:transfer}) &
  $\dagger$\,\texttt{models/} \texttt{ppo\_on\_pemirl/} \texttt{history.json}
  (seed $0$); per-seed histories in full repository \\
Effect sizes (Tab.~\ref{tab:eta}) &
  $\dagger$\,\texttt{results/reference/} \texttt{eta\_squared.json} \\
Feature hiding (Tab.~\ref{tab:hidden}) &
  $\dagger$\,\texttt{runs/eval/} \texttt{*\_hidden*.json} \\
Info-max variants &
  $\dagger$\,\texttt{runs/eval/} \texttt{robustness\_} \texttt{info10.json} \\
Landscape comparison &
  $\dagger$\,\texttt{runs/eval/} \texttt{*canonical\_airl*},
  \texttt{*gail*}, \texttt{*bc*}, \texttt{*seq\_*.json} \\
Context probes (Tab.~\ref{tab:probes}) &
  $\dagger$\,\texttt{models/pemirl/} \texttt{pemirl.pt};
  probe script in full repository \\
Stabilization recipe &
  $\dagger$\,\texttt{configs/pemirl.yaml}, with inline rationale \\
\midrule
$F$-ratios (Tab.~\ref{tab:decomp}) &
  descriptor script \texttt{02\_} \texttt{heterogeneity.py};
  per-voyage descriptors withheld as AIS derivatives \\
Second region &
  regional configs \texttt{configs/} \texttt{gsl*.yaml} and dataset builder \\
Semi-synthetic control (Tab.~\ref{tab:synth}) &
  generation, training, and probe scripts (full repository) \\
Original implementation &
  adapter and export scripts (full repository) \\
\bottomrule
\end{tabular}
\renewcommand{\arraystretch}{1}
\end{table}

The context-conditioned discriminator, bi-LSTM posterior encoder, info-max objective, and training loop are adapted from the PEMIRL implementation of \citet{MHAIRL} (\texttt{model/pemirl\_airl.py} and \texttt{model/context\_net.py}), modified for discrete, action-masked, goal-conditioned graph MDPs. Core loss functions and gradient penalty conventions are preserved, with stabilization modifications described in Section B. The replication analysis in Section I uses the original PEMIRL code release from \citet{yu2019meta}. Implementations of MCE-IRL \cite{ziebart2010modeling}, AIRL \cite{fu2018learning}, GAIL \cite{ho2016generative}, PPO \cite{schulman2017proximal}, and baseline models were developed from published specifications.

\section{I. Faithful Replication of the Original Implementation}
\label{app:original}

The original PEMIRL codebase \cite{yu2019meta} uses TensorFlow 1 and targets continuous control with fixed-length trajectories. Replicating this implementation required four interface adaptations:
\begin{enumerate}
\item Demonstration formatting: Training voyages ($1{,}000$) were converted to fixed-length trajectories of $100$ steps via truncation ($26.8\%$ of episodes) or goal padding.
\item Environment adapter: A graph replay environment matching state transitions, action masks, and features was constructed under the original environment interface.
\item Masked discrete policy: The categorical policy was modified with action masking, offsetting invalid logits rather than setting them to $-\infty$ due to entropy calculations.
\item Fidelity gating: Stabilization modifications were placed behind configuration flags and disabled to evaluate the published objective.
\end{enumerate}

Data exchanged between TensorFlow 1 and PyTorch frameworks were transferred via numeric arrays. Trained networks were re-implemented in NumPy and verified against TensorFlow outputs to numerical floating-point precision. State sequences were verified by replaying all demonstrations through the adapter. The original configuration uses a $3$-dimensional latent context, against $8$ dimensions in our implementation.

\paragraph{Results.}
Table~\ref{tab:original} reports the three seeds. The original
implementation obtains the best held-out likelihood of any model we
evaluate, and the best goal-reaching rate for decoded routes, but it does
not yield a transferable reward: a fresh PPO agent on its frozen reward
reaches $0.093$ of held-out goals on seed $0$, below the $0.089$
uniform-random baseline, and is highly seed-dependent
($0.35 \pm 0.24$ across seeds). Transferring the potential-shaped signal
$r(s,a,s') + \gamma V(s') - V(s)$ instead of the raw reward raises this to
$0.43 \pm 0.21$ (seed $0$: $0.156$ initial, $0.328$ peak, $0.200$
steady-state), indicating that usable signal is present but poorly scaled
in the unshaped reward.

Its likelihood advantage should not be read as evidence for latent
context. The original objective includes a one-shot imitation term on the
policy, and its per-decision likelihood ($-1.288 \pm 0.026$ micro) sits at
the pooled behavior-cloning anchor ($-1.26$) rather than above it. Its
latent is $3$-dimensional and its posterior underwent variance collapse
during training. As in Section~C, implementation and
objective choices move every metric substantially more than latent
context does.

\begin{table}[H]
\centering
\caption{Original PEMIRL implementation on the Arctic corpus, seeds
$\{0,1,2\}$, evaluated under our protocol. Transfer is the steady-state
goal-reaching rate of a fresh PPO agent on the frozen reward; shaped
transfer adds the potential terms $\gamma V(s') - V(s)$.}
\label{tab:original}
\setlength{\tabcolsep}{4pt}
\renewcommand{\arraystretch}{1.15}
\small
\begin{tabular}{lc}
\toprule
Metric & Original implementation \\
\midrule
LL micro $\uparrow$        & $\ms{-1.288}{0.026}$ \\
LL macro $\uparrow$        & $\ms{-1.141}{0.026}$ \\
Hausdorff (km) $\downarrow$ & $\ms{176.6}{15.4}$ \\
Reached goal $\uparrow$     & $\ms{0.53}{0.02}$ \\
FEE $\downarrow$            & $\ms{356}{15}$ \\
Transfer, raw $\uparrow$    & $\ms{0.35}{0.24}$ \\
Transfer, shaped $\uparrow$ & $\ms{0.43}{0.21}$ \\
\bottomrule
\end{tabular}
\renewcommand{\arraystretch}{1}
\end{table}

\paragraph{Operational observations.}
A run counts as diverged if the discriminator loss reaches $0$ or exceeds
$10^{3}$, if the mean recovered reward exceeds $10^{3}$ in magnitude, if
any loss is NaN, or if the final test likelihood falls below the
masked-uniform baseline. No seed tripped these criteria at the end of
training, but three operational issues were observed:
\begin{itemize}
\item Reward magnitude drift: The reward output magnitude grew unconstrained (reaching order $10^3$).
\item Numerical loss spikes: One seed experienced an infinite discriminator loss for a single iteration before recovering.
\item Context variance collapse: The posterior context encoder exhibited variance collapse across training episodes.
\end{itemize}

Unbounded reward scaling is consistent with the poor transfer performance observed with frozen rewards from this implementation.

\section{J. Semi-Synthetic Diagnostics}
\label{app:synthetic}

We use semi-synthetic experiments to test whether the evaluation
pipeline can detect known latent variation. These results apply only to
the constructed settings and the per-decision log-likelihood metric.

\paragraph{Corridor preferences.}
We create two latent navigation styles by adding
$\pm\lambda$ times normalized latitude to the MCE-IRL reward. We test
$\lambda \in \{0, 0.25, 0.5, 1.0, 2.0\}$. Each setting contains about
960 episodes from 40 synthetic vessels, with 30 vessels for training
and 10 for testing. AIRL uses no latent context; PEMIRL uses an
8-dimensional context.

As $\lambda$ increases, the style effect size increases from
$\eta^2=0.003$ to $0.321$ (Table~\ref{tab:synth}). However, AIRL
outperforms PEMIRL at every dose. At $\lambda=2.0$, an oracle policy
with true style labels improves over pooled behavior cloning by only
$0.031$ nats per decision. Thus, corridor preferences create limited
headroom under this metric because they change decisions at relatively
few graph locations.

\begin{table}[H]
\centering
\caption{Semi-synthetic corridor-preference results. Values are
held-out per-decision log-likelihood; higher is better. Gap is PEMIRL
minus AIRL.}
\label{tab:synth}
\setlength{\tabcolsep}{5pt}
\renewcommand{\arraystretch}{1.15}
\small
\begin{tabular}{lcccc}
\toprule
Dose $\lambda$ & $\eta^2$ & AIRL & PEMIRL & Gap \\
\midrule
$0.00$ & $0.003$ & $\mathbf{-1.342}$ & $-1.435$ & $-0.093$ \\
$0.25$ & $0.039$ & $\mathbf{-1.308}$ & $-1.394$ & $-0.086$ \\
$0.50$ & $0.118$ & $\mathbf{-1.285}$ & $-1.447$ & $-0.162$ \\
$1.00$ & $0.234$ & $\mathbf{-1.364}$ & $-1.435$ & $-0.071$ \\
$2.00$ & $0.321$ & $\mathbf{-1.316}$ & $-1.420$ & $-0.104$ \\
\bottomrule
\end{tabular}
\renewcommand{\arraystretch}{1}
\end{table}

\paragraph{Directional habits.}
We also create latent styles by adding action bonuses to alternating
compass directions. At the largest dose, an oracle with true style
labels has $0.133$ nats of headroom over a shared policy. AIRL achieves
$-1.480$ nats per decision, while PEMIRL achieves $-1.600$. The PEMIRL
posterior does not separate the two styles: a style-classification
probe is at chance accuracy ($0.500$). Increasing the information
weight to $\eta=0.1$ improves PEMIRL by approximately $0.06$ nats, but
it remains below AIRL.

These experiments use one seed and ten test vessels. They show that,
in this MDP, the tested forms of latent variation provide limited
per-decision likelihood headroom, and the evaluated PEMIRL posterior
does not recover that available signal.

\putbib
\end{bibunit}


\begin{thebibliography}{33}
\providecommand{\natexlab}[1]{#1}

\bibitem[{Amodei et~al.(2016)Amodei, Olah, Steinhardt, Christiano, Schulman, and Man{\'e}}]{amodei2016concrete}
Amodei, D.; Olah, C.; Steinhardt, J.; Christiano, P.; Schulman, J.; and Man{\'e}, D. 2016.
\newblock Concrete Problems in {AI} Safety.
\newblock arXiv:1606.06565.

\bibitem[{Chen et~al.(2023)Chen, Tamboli, Lan, and Aggarwal}]{MHAIRL}
Chen, J.; Tamboli, D.; Lan, T.; and Aggarwal, V. 2023.
\newblock Multi-task Hierarchical Adversarial Inverse Reinforcement Learning.
\newblock In \emph{Proc. ICML}, volume 202 of \emph{PMLR}, 4895--4920. PMLR.

\bibitem[{Chen et~al.(2016)Chen, Duan, Houthooft, Schulman, Sutskever, and Abbeel}]{chen2016infogan}
Chen, X.; Duan, Y.; Houthooft, R.; Schulman, J.; Sutskever, I.; and Abbeel, P. 2016.
\newblock {InfoGAN}: Interpretable Representation Learning by Information Maximizing Generative Adversarial Nets.
\newblock In \emph{Adv. Neural Inf. Process. Syst.}, volume~29, 2172--2180.

\bibitem[{Finn, Abbeel, and Levine(2017)}]{finn2017model}
Finn, C.; Abbeel, P.; and Levine, S. 2017.
\newblock Model-Agnostic Meta-Learning for Fast Adaptation of Deep Networks.
\newblock In \emph{Proc. ICML}, volume~70 of \emph{PMLR}, 1126--1135. PMLR.

\bibitem[{Finn, Levine, and Abbeel(2016)}]{finn2016guided}
Finn, C.; Levine, S.; and Abbeel, P. 2016.
\newblock Guided Cost Learning: Deep Inverse Optimal Control via Policy Optimization.
\newblock In \emph{Proc. ICML}, volume~48 of \emph{PMLR}, 49--58. PMLR.

\bibitem[{Fu, Luo, and Levine(2018)}]{fu2018learning}
Fu, J.; Luo, K.; and Levine, S. 2018.
\newblock Learning Robust Rewards with Adversarial Inverse Reinforcement Learning.
\newblock In \emph{Proc. ICLR}.

\bibitem[{Heckman(1981)}]{heckman1981}
Heckman, J.~J. 1981.
\newblock Heterogeneity and State Dependence.
\newblock In Rosen, S., ed., \emph{Studies in Labor Markets}, chapter~3, 91--140. Chicago, IL: Univ. Chicago Press.

\bibitem[{Hersbach et~al.(2020)Hersbach, Bell, Berrisford, Hirahara, Hor{\'a}nyi, Mu{\~n}oz-Sabater, Nicolas, Peubey, Radu, Schepers, Simmons, Soci, Abdalla, Abellan, Balsamo, Bechtold, Biavati, Bidlot, Bonavita, {De Chiara}, Dahlgren, Dee, Diamantakis, Dragani, Flemming, Forbes, Fuentes, Geer, Haimberger, Healy, Hogan, H{\'o}lm, Janiskov{\'a}, Keeley, Laloyaux, Lopez, Lupu, Radnoti, {de Rosnay}, Rozum, Vamborg, Villaume, and Th{\'e}paut}]{hersbach2020era5}
Hersbach, H.; Bell, B.; Berrisford, P.; Hirahara, S.; Hor{\'a}nyi, A.; Mu{\~n}oz-Sabater, J.; Nicolas, J.; Peubey, C.; Radu, R.; Schepers, D.; Simmons, A.; Soci, C.; Abdalla, S.; Abellan, X.; Balsamo, G.; Bechtold, P.; Biavati, G.; Bidlot, J.-R.; Bonavita, M.; {De Chiara}, G.; Dahlgren, P.; Dee, D.; Diamantakis, M.; Dragani, R.; Flemming, J.; Forbes, R.; Fuentes, M.; Geer, A.; Haimberger, L.; Healy, S.; Hogan, R.~J.; H{\'o}lm, E.; Janiskov{\'a}, M.; Keeley, S.; Laloyaux, P.; Lopez, P.; Lupu, C.; Radnoti, G.; {de Rosnay}, P.; Rozum, I.; Vamborg, F.; Villaume, S.; and Th{\'e}paut, J.-N. 2020.
\newblock The {ERA5} Global Reanalysis.
\newblock \emph{Q. J. R. Meteorol. Soc.}, 146(730): 1999--2049.

\bibitem[{Ho and Ermon(2016)}]{ho2016generative}
Ho, J.; and Ermon, S. 2016.
\newblock Generative Adversarial Imitation Learning.
\newblock In \emph{Adv. Neural Inf. Process. Syst.}, volume~29, 4565--4573.

\bibitem[{Howard and Hoffman(2018)}]{howard2018}
Howard, M.~C.; and Hoffman, M.~E. 2018.
\newblock Variable-Centered, Person-Centered, and Person-Specific Approaches: Where Theory Meets the Method.
\newblock \emph{Organ. Res. Methods}, 21(4): 846--876.

\bibitem[{Huang and Onta{\~n}{\'o}n(2022)}]{huang2020closer}
Huang, S.; and Onta{\~n}{\'o}n, S. 2022.
\newblock A Closer Look at Invalid Action Masking in Policy Gradient Algorithms.
\newblock In \emph{Proc. FLAIRS}, volume~35.

\bibitem[{Lakens(2013)}]{lakens2013calculating}
Lakens, D. 2013.
\newblock Calculating and Reporting Effect Sizes to Facilitate Cumulative Science: A Practical Primer for t-Tests and {ANOVA}s.
\newblock \emph{Front. Psychol.}, 4: 863.

\bibitem[{Laursen and Hoff(2006)}]{laursen2006}
Laursen, B.; and Hoff, E. 2006.
\newblock Person-Centered and Variable-Centered Approaches to Longitudinal Data.
\newblock \emph{Merrill-Palmer Q.}, 52(3): 377--389.

\bibitem[{Li, Song, and Ermon(2017)}]{li2017infogail}
Li, Y.; Song, J.; and Ermon, S. 2017.
\newblock {InfoGAIL}: Interpretable Imitation Learning from Visual Demonstrations.
\newblock In \emph{Adv. Neural Inf. Process. Syst.}, volume~30, 3812--3822.

\bibitem[{Liu, Zhu, and Zhang(2022)}]{liu2022goal}
Liu, M.; Zhu, M.; and Zhang, W. 2022.
\newblock Goal-Conditioned Reinforcement Learning: Problems and Solutions.
\newblock In \emph{Proc. IJCAI}, 5502--5511. IJCAI.

\bibitem[{Mannering, Shankar, and Bhat(2016)}]{mannering2016}
Mannering, F.~L.; Shankar, V.; and Bhat, C.~R. 2016.
\newblock Unobserved Heterogeneity and the Statistical Analysis of Highway Accident Data.
\newblock \emph{Anal. Methods Accid. Res.}, 11: 1--16.

\bibitem[{Nguyen and Fablet(2024)}]{nguyen2024traisformer}
Nguyen, D.; and Fablet, R. 2024.
\newblock A Transformer Network With Sparse Augmented Data Representation and Cross Entropy Loss for {AIS}-Based Vessel Trajectory Prediction.
\newblock \emph{IEEE Access}, 12: 21596--21609.

\bibitem[{Orsini et~al.(2021)Orsini, Raichuk, Hussenot, Vincent, Dadashi, Girgin, Geist, Bachem, Pietquin, and Andrychowicz}]{orsini2021matters}
Orsini, M.; Raichuk, A.; Hussenot, L.; Vincent, D.; Dadashi, R.; Girgin, S.; Geist, M.; Bachem, O.; Pietquin, O.; and Andrychowicz, M. 2021.
\newblock What Matters for Adversarial Imitation Learning?
\newblock In \emph{Adv. Neural Inf. Process. Syst.}, volume~34, 14656--14668.

\bibitem[{Osa et~al.(2018)Osa, Pajarinen, Neumann, Bagnell, Abbeel, and Peters}]{osa2018algorithmic}
Osa, T.; Pajarinen, J.; Neumann, G.; Bagnell, J.~A.; Abbeel, P.; and Peters, J. 2018.
\newblock An Algorithmic Perspective on Imitation Learning.
\newblock \emph{Found. Trends Robot.}, 7(1--2): 1--179.

\bibitem[{Rakelly et~al.(2019)Rakelly, Zhou, Finn, Levine, and Quillen}]{rakelly2019efficient}
Rakelly, K.; Zhou, A.; Finn, C.; Levine, S.; and Quillen, D. 2019.
\newblock Efficient Off-Policy Meta-Reinforcement Learning via Probabilistic Context Variables.
\newblock In \emph{Proc. ICML}, volume~97 of \emph{PMLR}, 5331--5340. PMLR.

\bibitem[{Schulman et~al.(2017)Schulman, Wolski, Dhariwal, Radford, and Klimov}]{schulman2017proximal}
Schulman, J.; Wolski, F.; Dhariwal, P.; Radford, A.; and Klimov, O. 2017.
\newblock Proximal Policy Optimization Algorithms.
\newblock arXiv:1707.06347.

\bibitem[{Shu et~al.(2026)Shu, Xu, Cui, Xiao, Song, Li, and Yang}]{shu2026integrating}
Shu, Y.; Xu, W.; Cui, H.; Xiao, J.; Song, L.; Li, H.; and Yang, Z. 2026.
\newblock Integrating Spatio-Temporal Analysis for Assessing the Effectiveness of {POLARIS} in Arctic Shipping Traffic.
\newblock \emph{Transp. Policy}, 181: 104096.

\bibitem[{Skalse et~al.(2022)Skalse, Howe, Krasheninnikov, and Krueger}]{skalse2022defining}
Skalse, J.; Howe, N.; Krasheninnikov, D.; and Krueger, D. 2022.
\newblock Defining and Characterizing Reward Gaming.
\newblock In \emph{Adv. Neural Inf. Process. Syst.}, volume~35, 9460--9471.

\bibitem[{Spadon et~al.(2025)Spadon, Song, Vaidheeswaran, Alam, Goerlandt, and Pelot}]{spadon2025modeling}
Spadon, G.; Song, R.; Vaidheeswaran, V.; Alam, M.~M.; Goerlandt, F.; and Pelot, R. 2025.
\newblock Modeling Maritime Transportation Behavior Using {AIS} Trajectories and {Markovian} Processes in the Gulf of St. Lawrence.
\newblock In \emph{Proc. IEEE Big Data}, 5314--5323. IEEE.

\bibitem[{Vaidheeswaran et~al.(2025)Vaidheeswaran, Jayakody, Mulay, Lo, Alam, and Spadon}]{vaidheeswaran2025goal}
Vaidheeswaran, V.; Jayakody, D.; Mulay, S.; Lo, A.; Alam, M.~M.; and Spadon, G. 2025.
\newblock Goal-Conditioned Reinforcement Learning for Data-Driven Maritime Navigation.
\newblock In \emph{Proc. IEEE Big Data}, 1194--1203. IEEE.

\bibitem[{Wang et~al.(2019)Wang, Wu, Zhao, Peng, and Lin}]{wang2019empowering}
Wang, J.; Wu, N.; Zhao, W.~X.; Peng, F.; and Lin, X. 2019.
\newblock Empowering {A*} Search Algorithms with Neural Networks for Personalized Route Recommendation.
\newblock In \emph{Proc. ACM SIGKDD}, 539--547. ACM.

\bibitem[{Xu et~al.(2019)Xu, Ratner, Dragan, Levine, and Finn}]{xu2019learning}
Xu, K.; Ratner, E.; Dragan, A.; Levine, S.; and Finn, C. 2019.
\newblock Learning a Prior over Intent via Meta-Inverse Reinforcement Learning.
\newblock In \emph{Proc. ICML}, volume~97 of \emph{PMLR}, 6952--6962. PMLR.

\bibitem[{Yu et~al.(2019)Yu, Yu, Finn, and Ermon}]{yu2019meta}
Yu, L.; Yu, T.; Finn, C.; and Ermon, S. 2019.
\newblock Meta-Inverse Reinforcement Learning with Probabilistic Context Variables.
\newblock In \emph{Adv. Neural Inf. Process. Syst.}, volume~32, 11772--11783.

\bibitem[{Zhang et~al.(2022)Zhang, Zhang, Zhang, Zhang, and Mao}]{zhang2022three}
Zhang, C.; Zhang, D.; Zhang, M.; Zhang, J.; and Mao, W. 2022.
\newblock A Three-Dimensional Ant Colony Algorithm for Multi-Objective Ice Routing of a Ship in the Arctic Area.
\newblock \emph{Ocean Eng.}, 266: 113241.

\bibitem[{Zheng(2015)}]{zheng2015trajectory}
Zheng, Y. 2015.
\newblock Trajectory Data Mining: An Overview.
\newblock \emph{ACM Trans. Intell. Syst. Technol.}, 6(3): 29:1--29:41.

\bibitem[{Ziebart, Bagnell, and Dey(2010)}]{ziebart2010modeling}
Ziebart, B.~D.; Bagnell, J.~A.; and Dey, A.~K. 2010.
\newblock Modeling Interaction via the Principle of Maximum Causal Entropy.
\newblock In \emph{Proc. ICML}, 1255--1262.

\bibitem[{Ziebart et~al.(2008)Ziebart, Maas, Bagnell, and Dey}]{ziebart2008maximum}
Ziebart, B.~D.; Maas, A.~L.; Bagnell, J.~A.; and Dey, A.~K. 2008.
\newblock Maximum Entropy Inverse Reinforcement Learning.
\newblock In \emph{Proc. AAAI}, 1433--1438.

\bibitem[{Zuo et~al.(2019)Zuo, Balmaseda, Tietsche, Mogensen, and Mayer}]{zuo2019oras5}
Zuo, H.; Balmaseda, M.~A.; Tietsche, S.; Mogensen, K.; and Mayer, M. 2019.
\newblock The {ECMWF} Operational Ensemble Reanalysis-Analysis System for Ocean and Sea Ice: A Description of the System and Assessment.
\newblock \emph{Ocean Sci.}, 15(3): 779--808.

\end{thebibliography}


\begin{thebibliography}{7}
\providecommand{\natexlab}[1]{#1}

\bibitem[{Chen et~al.(2023)Chen, Tamboli, Lan, and Aggarwal}]{MHAIRL}
Chen, J.; Tamboli, D.; Lan, T.; and Aggarwal, V. 2023.
\newblock Multi-task Hierarchical Adversarial Inverse Reinforcement Learning.
\newblock In \emph{Proc. ICML}, volume 202 of \emph{PMLR}, 4895--4920. PMLR.

\bibitem[{Fu, Luo, and Levine(2018)}]{fu2018learning}
Fu, J.; Luo, K.; and Levine, S. 2018.
\newblock Learning Robust Rewards with Adversarial Inverse Reinforcement Learning.
\newblock In \emph{Proc. ICLR}.

\bibitem[{Ho and Ermon(2016)}]{ho2016generative}
Ho, J.; and Ermon, S. 2016.
\newblock Generative Adversarial Imitation Learning.
\newblock In \emph{Adv. Neural Inf. Process. Syst.}, volume~29, 4565--4573.

\bibitem[{Olejnik and Algina(2003)}]{olejnik2003generalized}
Olejnik, S.; and Algina, J. 2003.
\newblock Generalized Eta and Omega Squared Statistics: Measures of Effect Size for Some Common Research Designs.
\newblock \emph{Psychol. Methods}, 8(4): 434--447.

\bibitem[{Schulman et~al.(2017)Schulman, Wolski, Dhariwal, Radford, and Klimov}]{schulman2017proximal}
Schulman, J.; Wolski, F.; Dhariwal, P.; Radford, A.; and Klimov, O. 2017.
\newblock Proximal Policy Optimization Algorithms.
\newblock arXiv:1707.06347.

\bibitem[{Yu et~al.(2019)Yu, Yu, Finn, and Ermon}]{yu2019meta}
Yu, L.; Yu, T.; Finn, C.; and Ermon, S. 2019.
\newblock Meta-Inverse Reinforcement Learning with Probabilistic Context Variables.
\newblock In \emph{Adv. Neural Inf. Process. Syst.}, volume~32, 11772--11783.

\bibitem[{Ziebart, Bagnell, and Dey(2010)}]{ziebart2010modeling}
Ziebart, B.~D.; Bagnell, J.~A.; and Dey, A.~K. 2010.
\newblock Modeling Interaction via the Principle of Maximum Causal Entropy.
\newblock In \emph{Proc. ICML}, 1255--1262.

\end{thebibliography}
\end{document}